# Gaussian Processes with Context-Supported Priors for Active Object Localization

Anthony D. Rhodes[1], Jordan Witte[2], Melanie Mitchell[2,3], Bruno Jedynak[1]
Portland State University: [1]Department of Mathematics and Statistics, [2]Computer Science; [3]Santa Fe Institute

## Abstract

*We devise an algorithm using a Bayesian optimization framework in conjunction with contextual visual data for the efficient localization of objects in still images. Recent research has demonstrated substantial progress in object localization and related tasks for computer vision. However, many current state-of-the-art object localization procedures still suffer from inaccuracy and inefficiency, in addition to failing to provide a principled and interpretable system amenable to high-level vision tasks. We address these issues with the current research.*

*Our method encompasses an active search procedure that uses contextual data to generate initial bounding-box proposals for a target object. We train a convolutional neural network to approximate an offset distance from the target object. Next, we use a Gaussian Process to model this offset response signal over the search space of the target. We then employ a Bayesian active search for accurate localization of the target.*

*In experiments, we compare our approach to a state-of-the-art bounding-box regression method for a challenging pedestrian localization task. Our method exhibits a substantial improvement over this baseline regression method.*

## 1. Introduction

Precise object localization remains an enduring, open challenge in computer vision. For example, fine-grained pedestrian localization in images is an active area of research with rich application potential [42]. More generally, accurate object localization is a vital task for many real-word applications of computer vision including: autonomous driving [12], cancer detection [21], image captioning [29], scene recognition [10] and robotics [24]. Current benchmark approaches [32] in object localization commonly apply a form of semi-exhaustive search, requiring a high volume—oftentimes thousands—of potentially expensive function evaluations, such as classifications by a convolutional neural network (CNN). Because of their black box nature, these methods often lack interpretability and neglect to incorporate top-down information including contextual and scene attributes.

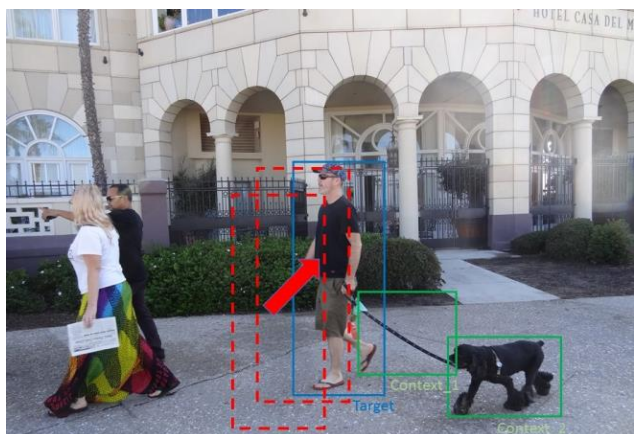

Figure 1: Idealization of localization process for pedestrian image using contextual data. Contextual data is shown in green; the ground-truth of the target is shown in blue, and target proposals are in red. Beginning with context-supported initial proposals, the GP-CL algorithm efficiently refines the localization process (All figures in this paper are best viewed in color.)

With [13][14], Girshick et al. achieved state-of-the-art performance on several object detection benchmarks using a "regions with convolutional neural networks" (R-CNN) approach. R-CNN comprises two phases: the region proposal generation and the proposal classification. Regional proposal generation renders rectangular regions of interest (ROIs) that are later classified by a deep CNN during proposal classification.

While the various R-CNN models perform well on general detection tasks, R-CNN-based approaches nonetheless suffer from at least (4) serious shortcomings and challenges: (1) the efficiency of the region proposal method, (2) the computational cost of evaluating the deep CNN, (3) localization accuracy and (4) the ability to successfully calibrate the R-CNN framework with top-down information, including context and feedback, in a principled, Bayesian manner.

We address each of these four areas by proposing a Bayesian optimization scheme in conjunction with contextual visual data for efficient object localization.

Our work provides the following contributions: (1) We demonstrate that CNN features computed from an object-proposal bounding box can be used to predict spatial offset from a target object. (2) We frame the localization process as an active search integrating top-down information in concert with a dynamic Bayesian optimization procedure requiring very few bounding-box proposals for high accuracy. (3) By rendering an active Bayesian search, our method can provide a principled and interpretable groundwork for more complex vision tasks, which we show explicitly through the incorporation of flexible context models. We compare our approach with the bounding-box regression method used in R-CNN approaches through experiments that test efficiency and accuracy for a challenging localization task.

The subsequent sections give some background on related work, the details of our method and algorithm, experimental results, summary remarks, and considerations of future work.

## 2. Background and Related Work

Object localization is the task of locating an instance of a particular object category in an image, typically by specifying a tightly-cropped bounding box centered on the instance. An object proposal specifies a candidate bounding box, and an object proposal is said to be a correct localization if it sufficiently overlaps a human-labeled "ground truth" bounding box for the given object. In the computer vision literature, overlap is measured via the intersection over union (IOU) of the two bounding boxes, and the threshold for successful localization is typically set to 0.5 [11]. In the literature, the "object localization" task is to locate one instance of an object category, whereas "object detection" focuses on locating all instances of a category in a given image.

For humans, recognizing a visual situation—and localizing its components—is an *active process* that unfolds over time, in which prior knowledge interacts with visual information as it is perceived to guide subsequent eye movements. This interaction enables a human viewer to very quickly locate relevant aspects of the situation [27].

Our method supports this more human-like approach of active object localization (e.g., [7], [15], [23]), in which a search for objects likewise unfolds over a series of time steps. At each time step the system uses information gained in previous time steps to decide where to search.

More recent variants of R-CNN, including, notably, Faster R-CNN [32], have attempted in the main to improve the efficiency of the core R-CNN pipeline by refining either the region proposal generation stage or the proposal classification stage of the localization algorithm. Faster R-CNN trains a region-proposal network (RPN) that shares full-image convolutional features with the detection network used in Fast R-CNN [13] to simultaneously predict object bounds and objectness scores. Other related methods (e.g., [18], [36]), attempt to simplify the CNN structure to improve computation time. Despite offering improvements, these methods still require considerable computing power [20].

Setting aside computational efficiency concerns, achieving accurate localization results is often an additional challenge in the R-CNN framework [43]. In particular, Hoiem et al. [17] show that inaccurate or "misaligned" bounding-boxes (i.e., boxes with a small IOU or intersection over union: $0.05 < IOU < 0.5$) exacerbate localization error for R-CNN. As such, R-CNN models are critically reliant on high-quality (i.e., $IOU > 0.5$) initial proposals; when no such proposals are present, R-CNN can render much weaker results [42]. We use a context-situation model, incorporating top-down, "situational" information to efficiently generate region proposals and then incorporate a Bayesian optimization scheme to further refine these proposals for accurate localization. The various R-CNN models all use category-specific "bounding-box regression" (BB-R) models to refine object proposals made by the system. In experiments, we compare our results against the BB-R models used by R-CNN for localization.

As an additional innovation, and in contrast to using the CNN as a discriminative object detector, we use features computed by a pretrained CNN to provide a localization "signal." We show that this signal (a function of the normalized offset distance of a bounding-box from the target ground-truth object) can be used effectively in a Bayesian optimization setting to quickly localize a target object.

The work of Zhang et al. [43] provides an extension of R-CNN that relates closely to the present work due to its use of Bayesian optimization. Despite this similarity, our work differs significantly in several important ways. Zhang et al., for instance, train their classifier as an object detector, whereas we instead train an offset-prediction signal. Furthermore, where Zhang et al. demonstrate a marginal improvement over baseline R-CNN on localization tasks, our method is fine-tuned for refining object proposals to guide an active localization procedure, particularly in the case of only marginally accurate initial proposals.

Context is described in terms of information that is necessary to characterize a visual situation. Recently, contextual information has been identified to improve several vectors of analysis in computer vision, including localization [39]. Indeed, the effective use of context is critical for future A.I. systems that aim to exhibit more comprehensive capabilities, including scene and situation "understanding" [30]. Nonetheless, many current systems disregard the use of context entirely, and its apposite use in vision tasks remains an open question.

Torralba and Murphy [25] incorporate global contextual features to learn context priors for object recognition. [26] frame localization as a MDP and apply unary and binary object contextual features to improve the search for a target object. Another successful use of context for localization includes [1] for which the class-specific search algorithm learns a strategy to localize objects by sequentially evaluating windows, based on statistical relation between the position and appearance of windows in the training images to their relative position with respect to the ground-truth. See also: [16], [6], [4], [28].

In the present work, related to [1], we learn contextual priors that model target object location and size. Together, we call the set of contextual priors a "context-situation model". [30] show that contextual information learned from situation-specific images can be successfully leveraged to improve localization. Using known contextual data from situation-specific images, we generate initial target proposals and then actively execute the search process using a Bayesian methodology – in this way the information gleaned from the prior can be weighed *actively* against evidence collected during the localization procedure.

## 3. Gaussian Processes with Context-Supported Priors for Active Object Localization

Gaussian Processes used in conjunction with a Bayesian optimization framework are frequently applied in domains for which it is either difficult or costly to directly evaluate an objective function. In the case of object detection and localization, it is computationally prohibitive to extract CNN features for numerous bounding-box proposals (this is why, for instance, Faster R-CNN utilizes shared convolutional features). There consequently exists a fundamental tension at the heart of any object localization paradigm: with each bounding box for which we extract CNN features, we gain useful knowledge that can be directly leveraged in the localization process, but each such piece of information comes at a price.

A Bayesian approach is well-suited for solving the problem of function optimization under these challenging circumstances. In the case of accurate object localization, we are attempting to minimize the spatial offset from a ground-truth bounding box (Figure 1). To do this, we train a model – described in Section 3.1 – to predict spatial offset of a proposal using CNN features extracted from the proposal. Once trained, the model output can be used to minimize the predicted offset. Ideally, this output is minimal when the proposal aligns with the actual ground-truth bounding box for the target object.

In our approach, we optimize a cheap approximation—the *surrogate* (also called the response surface) to the offset prediction—over the image space for efficiency. We give details of the realization of the surrogate function as a Gaussian process in Section 3.3.

Finally, after rendering this approximation, we determine where to sample next according to the principle of *maximum expected utility*. We identify utility using a dynamically defined *acquisition function* that strikes a balance between minimizing uncertainty and greedy optimization. This method is described in more detail in Section 3.4.

### 3.1 Training an Offset-Prediction Model

We train a model that predicts the normalized offset distance[1] from a target ground-truth object for a misaligned object proposal. The output of this model is the predicted distance of a proposal's center from the center of the target object, and the inverse of the output is the predicted proximity. We call the latter the "response signal." The higher the response signal, the closer the proposal is predicted to be to the target.

For each image in the training set, we generate a large number of image crops that are offset from the ground-truth pedestrian by a random amount. These randomized offset crops cover a wide range of IOU values (with respect to the ground-truth bounding box). These offset crops are also randomly scaled, so that the offset-prediction model can learn scale-invariance (with regard to bounding box size) for approximating offset distance. For each of the offset crops, we extracted CNN features using the pre-trained imagenet-vgg-f network in MatConvNet [44].

Using these features, we trained a ridge regression model mapping features to normalized offset distance from the ground-truth bounding box center. Next, we transformed this mapping in two steps using: (1) a scale transformation so that our feature-mapping scale corresponds to the bandwidth parameter used in the Gaussian process (see Section 3.3); and (2) a Gaussian-like transformation so that our prediction model renders an appropriate basin of attraction around the center of a target object that coheres with basic Gaussian process model assumptions. Note that in our regime, small offsets from the center of the target ground will yield (ideally) a maximum response signal. To improve the accuracy of our offset predictor, we average an ensemble of model outputs ranging over five different bounding-box scales.

The performance results of the offset-prediction model are plotted in Figure 2.

### 3.2 Context-Situation Learning

We define a context-situation model as a distribution of location and size parameters for a target object bounding-box, given various location and size parameters for a particular visual situation:

$$p(x_{target}, s_{target} | \{x_{context}, s_{context}\}_{1:C}) \quad (1)$$

---
[1] We use the Euclidean distance between the centers of two bounding boxes, scaled by the square root of the area of the image for the measure of "normalized offset distance."

where $x \in \mathbb{R}^2$ is the normalized bounding-box center, $s \in \mathbb{R}^2$ has components equal to the log bounding-box area-ratio (relative to the entire image) and log aspect-ratio, respectively; $C$ represents the number of known context objects.

In our experiments, we use a set of pedestrian images for our dataset (see section 4.1 for more detail) that comprise instances of a "dog-walking" visual situation; [17] showed that this learned context-situation facilitates improved object localization.

More specifically, this learned model consists of a set of probability distributions modeling the joint locations of the primary objects in the image as well as the joint area-ratios and aspect-ratios of bounding-boxes for these objects. These distributions capture the expected relationships among the objects with respect to location and size/shape of bounding-boxes. Naturally, these context-situation models can be extended and augmented as needed to improve compatibility and model expressiveness for a wide array of visual situations.

For simplicity and as a general proof of concept, we model context-situation as decoupled[2] size and shape MVN (multi-variate Normal) distributions. See Section 5 for comments regarding considerations of more robust density models for context-situation learning.

### 3.3 Gaussian Processes

We use a Gaussian Process (GP) to compute a surrogate function $f$ using observations $\{y\}$ of response signals from our prediction model: $y(x) = f_0(x) + \varepsilon$. (Recall that the signal $y$ is high when the input proposal is predicted to be close to the target object.) The surrogate function approximates $f_0$, the objective signal value for coordinates $x$ in the image space, with $\varepsilon$ connoting the irreducible error for the model.

GPs offer significant advantages over other general-purpose approaches in supervised learning settings due in part to their non-parametric structure, relative ease of computation and the extent to which they pair well with a Bayesian modeling regime. GPs have been applied recently with success in a rich variety of statistical inference domains, including [5], [41], [9].

More formally, we let $x_i \in \mathbb{R}^2$ be the $i$th observation from a dataset $D_{1:T} = \{x_{1:T}, y(x_{1:T})\}$ consisting of $T$ total pairs of object-proposal coordinates $x$ in the image space and response signals $y$, respectively. We wish to estimate the posterior distribution $p(f|D_{1:T})$ of the objective function given these data: $p(f|D_{1:T}) \propto p(D_{1:T}|f)p(f)$. This simple formula allows us to iteratively update the posterior over the signal as we acquire new data.

A GP for regression defines a distribution over functions with a joint Normality assumption. We denote $f$, the realization of the Gaussian process:

$$f \sim GP(m, k) \qquad (2)$$

Here the GP is fully specified by the mean $m$ and covariance $k$. A common kernel function that obeys suitable continuity characteristics for the GP realization is the squared-exponential kernel, which we use here:

$$k(x, x') = \sigma_f^2 exp\left[-\frac{1}{2l^2}\|x - x'\|^2\right] + \sigma_\varepsilon^2 \delta_{xx'}, \quad (3)$$

where $\sigma_f^2$ is the variance of the GP realization, which we set heuristically; $\sigma_\varepsilon^2$ is the variance of the $\varepsilon$ parameter that we estimate empirically; and $\delta_{xx'}$ is the *Kroenecker delta function* which is equal to 1 if and only if $x = x'$ and is equal to zero otherwise. GPs are particularly sensitive to the choice of the length-scale/bandwidth parameter $l$, which we optimize with grid search for the reduced log marginal likelihood (see [18] for additional details).

The posterior predictive of the surrogate function for a new datum $x_*$ is given by [3]:

$$p(f_*|x_*, X, y) = N(f_*|k_*^T K_\sigma^{-1} y, k_{**} - k_*^T K_\sigma^{-1} k_*) \quad (4)$$

where $X$ is the data matrix (all prior observations $x$), $k_* = [k(x_*, x_1), \ldots, k(x_*, x_T)], k_{**} = k(x_*, x_*)$
and $K_\sigma = K + \sigma_y^2 I_T$, where $K = k(x_i, x_j), 1 \leq i, j \leq T$.

For our algorithm, we compute posterior predictive updates using equation (4) in batch iterations (see Section 4.2). At each iteration, the realization of the GP is calculated over a grid of size $M$ corresponding with the image space domain of the object localization process. This grid size can be chosen to match a desired granularity/computational overhead tradeoff.

Considering equation (4) further, we note that posterior predictive updates entail a one-time (per iteration) inversion of the matrix $K_\sigma$, requiring $O(T^3)$ operations, where $T$ is the number of calls to the offset-prediction model. Naturally, choosing information-rich bounding-box proposals (see Section 3.4) will improve the efficiency of the localization process and thus keep $T$ reasonably small in general. To this end, we furthermore incorporate a "short memory" mechanism in our algorithm so that older proposal query values, which convey less information pertinent to the current localization search, are "forgotten" (see Section 4). For improved numerical stability, we apply a Cholesky decomposition prior to matrix inversion [31].

---

[2] By "decoupled" we mean that the location and size parameters are treated as independent densities, to minimize overfitting for photographic bias.

## 3.4 Bayesian Optimization for Active Search

In the regime of Bayesian optimization, acquisition functions are used to guide the search for the optimum of the surrogate approximating the true objective function. Intuitively, acquisition functions are defined in such a way that *high acquisition* indicates greater likelihood of an objective function optimum. Most commonly, acquisition functions encapsulate a data query experimental design that favors either regions of large signal response, large uncertainty, or a combination of both.

One can formally express the *utility* of a Bayesian optimization procedure with GP parameter θ, observations $\{y\}$, and acquisition function instantiated by $a(\xi)$ with design parameter $\xi \geq 0$, as the information gained when we update our prior belief $p(\theta|a(\xi))$ to the posterior, $p(\theta|y, a(\xi))$, after having acquired a new observation [3].

At each iteration of our algorithm, the acquisition function, defined below, is maximized to determine where to sample from the objective function (i.e., the response signal value) next. The acquisition function incorporates the mean and variance of the predictions over the image space to model the utility of sampling [3]. We then evaluate the objective function at these maximal points and the Gaussian process is updated appropriately. This procedure is iterated until the stopping condition is achieved.

A standard acquisition function used in applications of Bayesian optimization is the *Expected Improvement* (EI) function [37]. We define a dynamic variant of EI that we call *Confidence-EI* (CEI) that better accommodates our problem setting:

$$a_{CEI}(x,\xi) \triangleq \begin{cases} (\mu(x) - f(x^+) - \xi)\Phi(Z) + \sigma(x)\varphi(Z) \\ Z = \dfrac{\mu(x) - f(x^+) - \xi}{\sigma(x)} \end{cases} \quad (5)$$

In equation (5), $f(x^+)$ represents the incumbent maximum of the surrogate function, $\mu(x)$ is the mean of the surrogate at the input point $x$ in the image space, $\sigma(x) > 0$ is the standard deviation of the surrogate at the input; $\varphi(\cdot)$ and $\Phi(\cdot)$ are the *pdf* and *cdf* of the Gaussian distribution, respectively; and $\xi$ is the dynamically-assigned design parameter. The design parameter controls the exploration-exploitation tradeoff for the Bayesian optimization procedure; if, for instance, we set $\xi = 0$, then EI performs greedily.

For our algorithm, we let $\xi$ vary over the course of localization run by defining it as a function of a per-iteration *total confidence score*. Lizotte [22] showed that varying the design parameter can improve performance for Bayesian optimization. With each iteration of localization, we set the current total confidence value equal to the median of the response signal for the current batch of bounding-box proposals. In this way, high confidence disposes the search to be greedy and conversely low confidence encourages exploration.

## 4. Algorithm and Experimental Results

### 4.1 Dataset

Following [30] and [33], in the current study we use a dataset consisting of single pedestrian instances from the Portland State Dog-Walking Images for our proof of concept and comparative experiments [45]. This dataset contains 460 high-resolution annotated photographs, taken in a variety of locations. Each image is an instance of a "Dog-Walking" visual situation in a natural setting containing visible pedestrians. Quinn et al. [30] used this dataset to demonstrate the utility of applying prior situation knowledge and active, context-directed search in a structured visual situation for efficient object localization. These images represent a challenging benchmark for pedestrian localization, due to its high degree of variability and large image resolution.

### 4.2 GP-CL Algorithm

Below we present details of the Gaussian Process Context Localization (GP-CL) algorithm. To begin, we randomly set aside 400 images from our dataset for training and 60 for testing. We train the prediction model, *y*, using features computed by the pre-trained imagenet-vgg-f network in MatConvNet [44]. The features we use are from the last fully-connected layer, which yields feature vectors of dimension 4096. During training, we generated 100k offset crops of pedestrians from the training images.

For our context-situation model, we fit joint log-Normal distributions: $p(\cdot)_x$, $p(\cdot)_s$, for target object location and size, respectively, conditioned on the known location and size of the contextual objects consisting of dog and leash. For our purposes, we assume that these context objects are "perfectly" localized – only to prove that contextual data in concert with a Gaussian Process-directed search yields very efficient and precise localizations in general. To this end, [30] showed that "imperfect" contextual data is still viable for use in a refined localization procedure; in addition, the Bayesian nature of the present work effectively mitigates the influence of poor initial proposals. Note that because GP-CL algorithm employs a 2-d realization of a Gaussian Process for object location, $p(\cdot)_x$ serves as a prior for target location and $p(\cdot)_s$ functions as a prior for target size with regard to the initial proposal set. Thereafter, the Bayesian optimization procedure generates subsequent location proposals, while the size proposals continue to be drawn from the context-situation model for $p(\cdot)_s$.

We optimize the hyperparameter θ for the Gaussian process using grid search. The design parameter $\xi$ is set as a function of the per-step total. Lastly, we set the size of the GP realization, $M = 500^2$ (i.e., the realization occurs over a 500x500 grid). We found that this size achieved a suitable balance between localization precision and computational overhead.

For GP-CL, we begin by generating a set of ($n_0 = 10$). initial bounding-box proposals from the learned context-situation model. We then use our trained off-set prediction model to compute response signal values for this proposal set, yielding $D_{proposal}^{(0)}$. At each subsequent step of the GP-CL algorithm we generate a GP realization using the proposal set (step 4). To find the next batch ($n = 5$) of proposals, we use the top-$n$ ranked points in the space, ranked using the CEI acquisition function defined in Section 3.4. We then augment the proposal set with this new batch of points and the previous generations of proposals specified by the GP$_{mem}$ parameter, which indicates the number of batches contained in the algorithm "memory" (steps 10 and 11). For our experiments, we set GP$_{mem}$= 3 with $T$ = 10, for a total of 50 proposals per execution of GP-CL.

---

**Algorithm: Gaussian Process Context Localization** (GP-CL)

**Input**: Image *I*, a set of *C* context objects, trained model *y* giving response signals, learned context-situation model $p(x_{target}, s_{target} | \cdot)$, $n_0$ initial bounding-box proposals for target object generated by the context-situation model, and corresponding response signal values: $D_{n_0} = \{(x_i, s_i), y(x_i, s_i)\}_{i=1}^{n_0}$, GP hyperparameters θ, size of GP realization space *M*, dynamic design parameter for Bayesian active search $\xi$, size of GP memory GP$_{mem}$ (as number of generations used), batch size *n*, number of iterations *T*, current set of bounding-box proposals and response signals $D_{proposal}^{(t)}$.

1: Compute $n_0$ initial bounding box proposals:
 $\{(x_i, s_i)\}_{i=1}^{n_0} \sim p(x_{target}, s_{target} | \cdot)$
2: $D_{proposal}^{(0)} \leftarrow D_{n_0}$
3: **for** $t = 1$ to $T$ **do**
4:  Compute $\mu(x)^{(t)}$ and $\sigma(x)^{(t)}$ for the GP realization
  $f_M^{(t)}$ of $D_{proposal}^{(t-1)}$ over grid of *M* points (Equation 4)
5:  **for** $i = 1$ to $n$ **do**
6:   $z_i = \underset{x}{\operatorname{argmax}}\, a_{CEI}\left(f_M^{(t)} \setminus \{z_j\}_{j=1}^{j=i-1}, \xi\right)$ (Equation 5)
7:   $sample: s_i \sim p(\cdot)_s$
8:   $p_i = (z_i, s_i)$
9:  **end for**
10:  $D^{(t)} \leftarrow \{(x_i, s_i), y(x_i, s_i)\}_{i=1}^{n}$
11:  $D_{proposal}^{(t)} \leftarrow \bigcup_{j=t-GP_{mem}}^{t} D^{(j)}$
12: **end for**
13: **Return** $\underset{x}{\operatorname{argmax}} \mu(x)^{(T)}$

---

## 4.3 Experimental Results

We evaluate the GP-CL algorithm described in Section 4.1 in comparison with the benchmark bounding-box regression model used in Faster R-CNN [32] for the task of pedestrian localization. Both the GP and bounding-box regression models were trained with 100k offset image crops taken from the test image set. For the bounding-box regression trials, the algorithm receives a randomized offset crop in the IOU range [0, 0.7], and then outputs a refined bounding box. In the case of GP-CL, the algorithm is initialized with a small set ($n_0 = 10$) of proposals drawn from the context-situation model; this likewise resulted in initial proposals in the range [0, 0.7]. Because of both the challenging nature of our dataset and various simplifying assumptions implicit in the context model we used, in a small number of cases the context-situation model produced erroneous initial proposals (e.g. proposals centered outside the test image). In these cases we initialized the proposals with a random offset value. The data transformations applied to produce the offset-prediction model described in Section 3.1 were determined heuristically

The median IOU over all the initial proposal bounding-boxes for the GP-CL experimental trials was 0.23. Our context data consisted of perfect localizations of dogs and leashes in the "dog-walking" visual situation with pedestrians. [30] showed that imperfect context-based priors are still effective for improving the efficiency of localization. Our method, furthermore, is general enough to incorporate a variety of contextual models to serve as priors for the GP. In the case of the absence of contextual data, our approach also serves very effectively as a proposal "refinement" procedure.

The output of the GP-CL algorithm is a single bounding-box, as in the case of the regression model. For each method, we compare the final bounding-box with the ground-truth for the target object. In total, we tested each method for 440 experimental trials, including multiple runs with different initializations on test images.

Girshick et al. [14] thresholded their training regime for localization with bounding-box regression at large bounding-box overlap (IOU ≥ 0.6). To comprehensively test our method against bounding-box regression (BB-R), we trained two distinct regression models: one with IOU thresholded for training at 0.6, as used with R-CNN, and one with IOU thresholded at 0.1.

Results for our experiments are summarized in Table 1 and Figure 4. We report the median and standard error (SE) for IOU difference (final – initial), the median relative IOU improvement (final – initial) / initial, the total percentage of the test data for which the method yielded an IOU improvement, in addition to the total percentage of test data for which the target was successfully localized (i.e., final IOU ≥ 0.5).

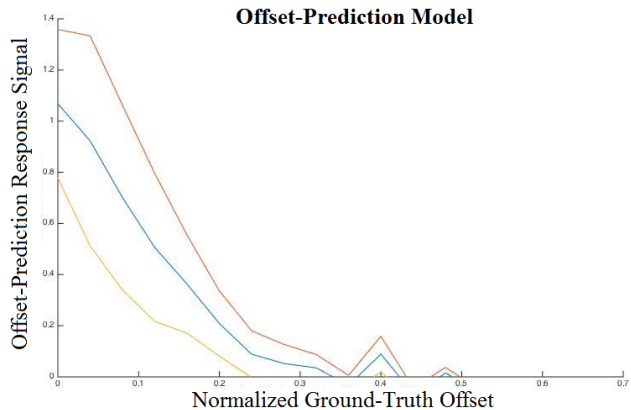

Figure 2: Performance of the offset-prediction model on test data ($n = 1000$ offset image crops). The mean (center curve) and +/−1 standard deviations (outer curves) are shown. As desired, the response signal yields a Gaussian-like peak around the center of the target object bounding-box (i.e., zero ground-truth offset). The bumps present in the range of values above 0.35 offset from the ground truth is indicative of noisy model outputs when offset crops contain no overlap with the target object. (Figure is best viewed in color.)

| Method | IOU Difference Median (SE) | Median Relative IOU Improvement | % of Test Set with IOU Improvement | % of Test Set Localized |
|---|---|---|---|---|
| BB-R (0.6) | .1065 (.004) | 32.35% | **93.86%** | 48.2% |
| BB-R (0.1) | .1034 (.009) | 29.0% | 71.1% | 44.1% |
| **GP-CL** | **.4938** (.012) | **134.7%** | 87.1% | **75.7%** |

Table 1: Summary statistics for the pedestrian localization task. BB-R (0.6) indicates the bounding-box regression model with training thresholded at initial IOU 0.6 and above; BB-R (0.1) denotes the bounding-box regression model with training thresholded at initial IOU 0.1 and above; GP-CL denotes Gaussian Process Context Localization. GP-CL can be seen to consistently outperform BB-R methods.

## 4.4 Discussion

Our experimental results are strongly favorable for the GP-CL algorithm. Using only a small number of total bounding box proposals (50) per trial, GP-CL performed comparably with BB-R for percentage of test images for which the IOU improved. In addition, GP-CL significantly outperformed BB-R for all other localization metrics, including the percentage of test set images achieving successful localization and the median relative IOU improvement.

During our experimental trials, we discovered a substantial disparity in performance for BB-R depending on the training regime. In general, BB-R (0.6), as used in R-CNN, yielded inferior localization results in general when compared to BB-R (0.1) (see Table 1). In particular, BB-R (0.1) was much stronger for low initial IOU values than BB-R (0.6). However, as initial IOU increased, localization results deteriorated starkly with BB-R (0.1) due to overfitting. For larger initial IOU values (e.g., IOU > 0.4), BB-R (0.1) yielded IOU improvement on only 22.1% of the experimental trials; when the IOU threshold was increased to 0.5 this IOU improvement percentage dropped even further to 13.0%. In contrast, GP-CL indicated no signs of deterioration in localization performance when given initial offset proposals with a large IOU. For separate test runs of 100 trials each, GP-CL achieved an IOU improvement on 97% of the trials (for median initial IOU > 0.4) and an IOU improvement on 99% of the trials (for median initial IOU > 0.5).

In addition to this strong experimental performance, GP-CL provides several broad methodological advantages over previous techniques, particularly in applications requiring fast and precise object localization. Most importantly, by working within a Bayesian framework, GP-CL is able to perform an efficient, active search by "learning" continuously from its response signal at each step of the algorithm. Because GP-CL renders both the mean and standard deviation for the predictive posterior, the GP-CL model maintains a measure of uncertainty that can be applied in systems as a potential (early) stopping condition when real-world resources are limited (e.g. robotics, video tracking using Kalman filters). As we show, a context model can be naturally and successfully integrated into the Gaussian Process framework.

## 5. Conclusion and Future Work

We have presented a novel technique for the challenging task of efficient object localization. Our method trains a predicted-offset model, demonstrating successfully the ability of CNN-based features to serve as the input for an object localization method. Using Bayesian optimization, we surpass the state-of-the-art regression method employed in R-CNN (and its extensions) for the localization of pedestrians in high-resolution still images with computational efficiency.

With future research, we plan to extend our approach to gradient-based GPs and massively scalable GPs, so that our model can directly incorporate bounding-box size parameters, as well as leverage additional sources of visual context for localization. More generally, we aim to apply these approaches to broader, "big data" and related high-dimensional problem regimes.

## Acknowledgements

This material is based upon work supported by the National Science Foundation under Grant Number IIS-1423651. Any opinions, findings, and conclusions or recommendations expressed in this material are those of the author sand do not necessarily reflect the views of the National Science Foundation.

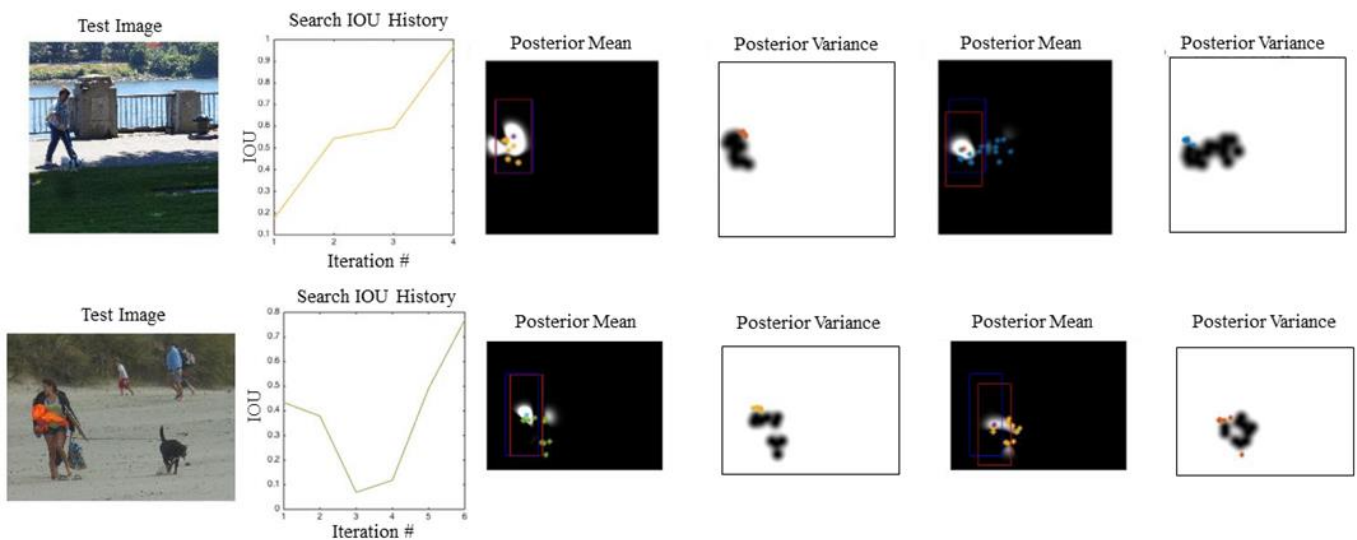

Figure 3: Examples of runs on two test images with the GP-CL algorithm. In each row the test image is shown on the far-left; the "search IOU history" is displayed in the second column, with the algorithm iteration number on the horizontal axis and IOU with the ground-truth target bounding box on the vertical axis. The remaining columns present the GP-CL response surface for the posterior mean and variance for target object location. In the first row, this pair of boxes reflect the third iteration of the algorithm and the last pair show the second iteration, respectively; in the second row, these pairs of boxes represent the sixth and fifth iterations of the algorithm, respectively. The red rectangle signifies the target object ground-truth bounding box, while the blue rectangle indicates the highest posterior mean response for the target object location at the current iteration. The colored dots in the "posterior mean" image show the sample batch for the current iteration; the colored dots in the "posterior variance" image indicate points with maximum CEI (confidence-expected improvement) scores following the current sampling batch. In each case localization occurs rapidly thus requiring a very small number of proposals.

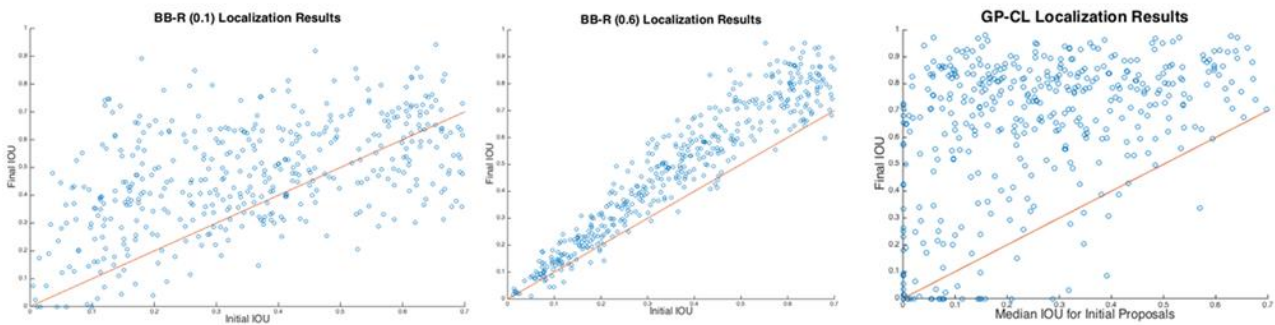

Figure 4: Graph of BB-R (0.6), BB-R (0.1) and GP-CL localization results for test images. The horizontal axis indicates the median IOU for the initial proposal bounding boxes, while the vertical axis designates the final IOU with the target object ground truth. The line depicted indicates "break-even" results. GP-CL reliably improves target object IOU for a broad range of initial IOU values.